\documentclass[11pt]{article}
\usepackage[final]{acl}
\usepackage{booktabs}
\usepackage[utf8]{inputenc}
\usepackage{booktabs} 
\usepackage{multirow} 
\usepackage{geometry} 
\usepackage{times}
\usepackage{latexsym}
\usepackage{amsmath}
\usepackage{makecell}
\usepackage[T1]{fontenc}
\usepackage{hyperref}
\usepackage{fontawesome5}

\usepackage[utf8]{inputenc}
\usepackage{relsize}
\usepackage{microtype}
\usepackage{graphicx}
\usepackage{colortbl}
\definecolor{bestcolor}{HTML}{FFF2C2} 
\definecolor{secondcolor}{HTML}{FFF8DE}   
\usepackage{inconsolata}
\usepackage[utf8]{inputenc}
\usepackage{booktabs} 
\usepackage{multirow} 
\usepackage{graphicx}
\usepackage{xcolor} 
\usepackage{amsfonts} 
\definecolor{myred}{HTML}{C00000}
\definecolor{mygreen}{RGB}{0, 150, 0}
\usepackage{amsmath}
\newcommand{\upred}[1]{$_{\color{myred}\uparrow #1}$}   
\newcommand{\downred}[1]{$_{\color{myred}\downarrow #1}$} 
\newcommand{\upgreen}[1]{$_{\color{mygreen}\uparrow #1}$} 
\newcommand{\downgreen}[1]{$_{\color{mygreen}\downarrow #1}$}

\usepackage[most]{tcolorbox}
\usepackage{xcolor}
\usepackage{listings}
\usepackage{tikz}
\usetikzlibrary{shadows}
\usepackage{algorithm}
\usepackage{algpseudocode}
\definecolor{boxPurple}{HTML}{8491E8} 
\definecolor{codebg}{HTML}{FFFFFF}
\usepackage{roboto-mono} 
\usepackage[T1]{fontenc}
\newcommand{\codefontsize}{\fontsize{8pt}{10.5pt}\selectfont}
\lstdefinelanguage{json}{
    basicstyle=\ttfamily\codefontsize, 
    columns=fullflexible,
    backgroundcolor=\color{codebg},    
    showstringspaces=false,
    commentstyle=\color{white},
    keywordstyle=\color{blue},
    stringstyle=\color{teal},
    breaklines=true,
    breakatwhitespace=true,
    frame=none, 
    literate=
     *{0}{{{\color{purple}0}}}{1}
      {1}{{{\color{purple}1}}}{1}
      {2}{{{\color{purple}2}}}{1}
      {3}{{{\color{purple}3}}}{1}
      {4}{{{\color{purple}4}}}{1}
      {5}{{{\color{purple}5}}}{1}
      {6}{{{\color{purple}6}}}{1}
      {7}{{{\color{purple}7}}}{1}
      {8}{{{\color{purple}8}}}{1}
      {9}{{{\color{purple}9}}}{1}
      {:}{{{\color{black}:}}}{1}
      {,}{{{\color{black},}}}{1}
      {\{}{{{\color{black}\{}}}{1}     
      {\}}{{{\color{black}\}}}}{1}     
      {[}{{{\color{black}[}}}{1}
      {]}{{{\color{black}]}}}{1},
}

\newtcblisting{nicejson}[2][]{
    enhanced,
    colback=codebg,        
    colframe=boxPurple,    
    colbacktitle=boxPurple,
    coltitle=white,        
    fonttitle=\Large\rmfamily,
    title={#2},            
    sharp corners,
    drop shadow={opacity=1, shadow xshift=1.5mm, shadow yshift=-1.5mm, color=boxPurple},
    listing only,
    listing options={language=json}, 
    #1
}
\title{\textit{Learning How to Remember:} A Meta-Cognitive Management Method for Structured and Transferable Agent Memory}

\author{
    Sirui Liang\textsuperscript{1,2,3}\thanks{\ \ Co-first authors, they contributed equally to this work.}, 
    Pengfei Cao\textsuperscript{1,2}\footnotemark[1], 
    Jian Zhao\textsuperscript{3,4}\thanks{\ \ Corresponding author.}, 
    \textbf{Wenhao Teng} \textsuperscript{5}, \\
    \textbf{Xiangwen Liao} \textsuperscript{6}, 
    \textbf{Jun Zhao}\textsuperscript{1,2}, 
    \textbf{Kang Liu}\textsuperscript{1,2}\footnotemark[2] \\
    \fontsize{10pt}{6pt}\selectfont
    $^1$Institute of Automation, CAS, $^2$University of Chinese Academy of Sciences, \\
    \fontsize{10pt}{6pt}\selectfont
    $^3$Zhongguancun Academy, $^4$Zhongguancun Institute of Artificial Intelligence, \\
    \fontsize{10pt}{6pt}\selectfont
    $^5$Department of Gastrointestinal Surgery, Fujian Provincial Cancer Hospital, \\
    \fontsize{10pt}{6pt}\selectfont
    $^6$College of Computer and Data Science, Fuzhou University \\
    \texttt{\small liangsirui2024@ia.ac.cn, jianzhao@zgci.ac.cn, \{pengfei.cao,kliu,jzhao\}@nlpr.ia.ac.cn}\\
    \faGithub\ \href{https://github.com/LiangThree/MCMA.git}{github.com/LiangThree/MCMA}
}

\begin{document}
\maketitle
\begin{abstract}
Large language model (LLM) agents increasingly rely on accumulated memory to solve long-horizon decision-making tasks. However, most existing approaches store memory in fixed representations and reuse it at a single or implicit level of abstraction, which limits generalization and often leads to negative transfer when distribution shift. This paper proposes the \textbf{M}eta-\textbf{C}ognitive \textbf{M}emory \textbf{A}bstraction method (\textbf{MCMA}), which treats memory abstraction as a learnable cognitive skill rather than a fixed design choice. MCMA decouples task execution from memory management by combining a frozen task model with a learned memory copilot. The memory copilot is trained using direct preference optimization, it determines how memories should be structured, abstracted, and reused. Memories are further organized into a hierarchy of abstraction levels, enabling selective reuse based on task similarity. When no memory is transferable, MCMA transfers the ability to abstract and manage memory by transferring the memory copilot. Experiments on ALFWorld, ScienceWorld, and BabyAI demonstrate substantial improvements in performance, out-of-distribution generalization, and cross-task transfer over several baselines.
\end{abstract}

\section{Introduction}

Large language model (LLM) agents have recently demonstrated strong performance beyond static question answering \cite{achiam2023gpt, bai2023qwen, chen2025swe, tao2024survey}, increasingly operating in long-horizon, interactive environments that require sustained decision making and environment feedback \cite{qiao2024agent, tan2025prospect, zhang2025survey, chhikara2025mem0}. In these settings, a key capability of agents is to accumulate, organize, and reuse memory, which is also known as \textbf{procedural memory} \cite{fang2025memp, cao2025remember}. Effectively reusing accumulated memories is essential for enabling agents to solve new tasks efficiently and operate robustly in complex, long-horizon environments \cite{song2024agentbank}.

\begin{figure}[t]
\centering
  \includegraphics[width=0.9\columnwidth]{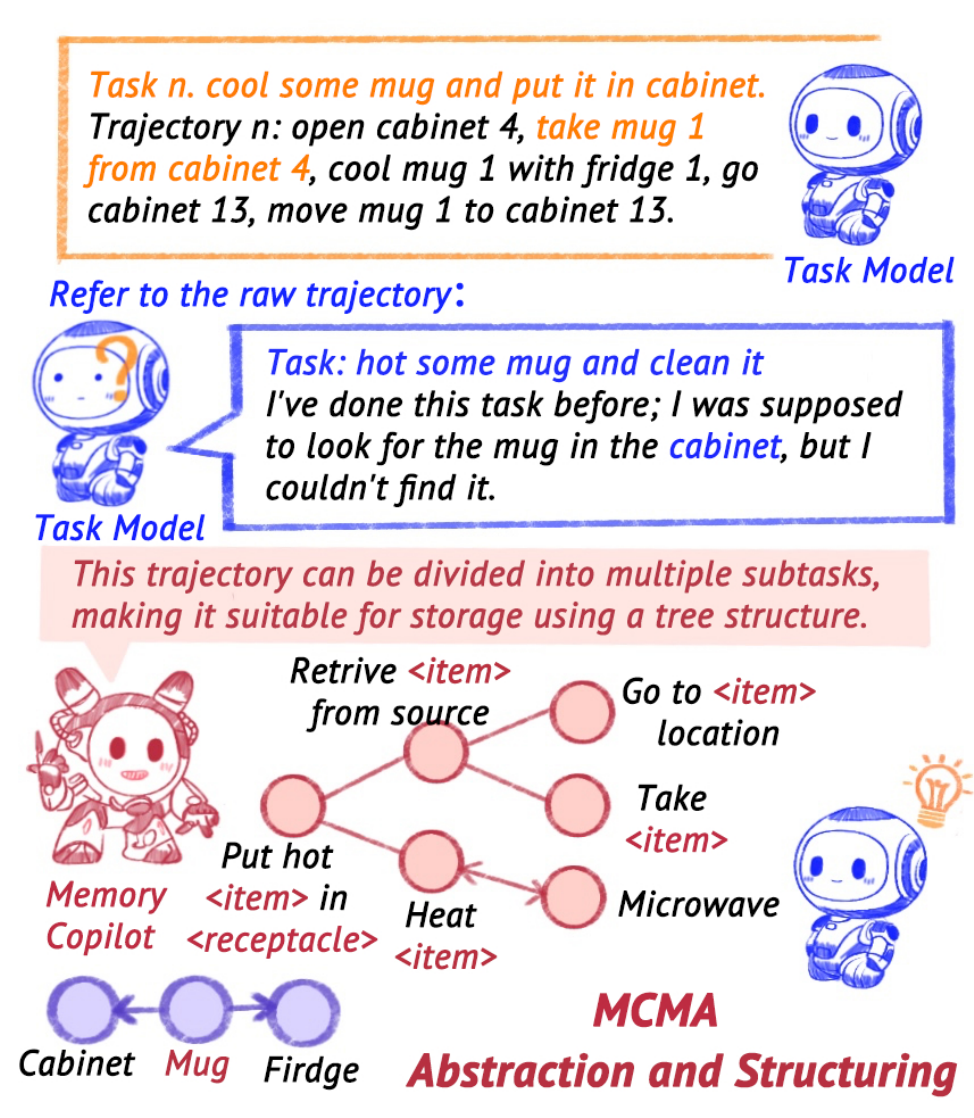}
  \vspace{-5pt}
  \caption{An example of how MCMA works.}
  \label{fig:mcma examples}
  \vspace{-10pt}
\end{figure}

Despite the growing use of memory in LLM-based agents, effective memory reuse remains challenging. Retrieval-based approaches recall trajectories \cite{zheng2023synapse}, sub-tasks \cite{kim2024rada}, or fine-grained execution steps \cite{zhou2024trad}, but often rely on surface similarity and fail under environment or task changes \cite{awm2024wang}. Summarization \cite{zhao2024expel}, abstraction \cite{awm2024wang}, and hierarchical methods \cite{ye2025h} mitigate noise by organizing memory at higher levels, yet still face an abstraction reuse dilemma. We believe this phenomenon is because fine-grained representations overfit to the seen environments, while overly abstract ones lack actionable guidance. Training-based approaches \cite{song2024agentbank, zeng2024agenttuning, wang2024wise} further internalize experience into model parameters to improve generalization, but tightly entangle memory with policy learning, limiting cross-task transfer and risking catastrophic forgetting under domain shifts \cite{chen2025internalizing}. Overall, the aforementioned methods rely on predefined memory representations and fixed or implicit abstraction levels. As a result, agents fail to learn reusable abstractions or adaptively select appropriate levels of memory abstraction for unseen tasks.

To address these challenges, this paper proposes the \textbf{M}eta-\textbf{C}ognitive \textbf{M}emory \textbf{A}bstraction method (\textbf{MCMA}), which treats memory abstraction as a learnable cognitive skill, rather than a static design choice. MCMA operates at a cognitive level by learning how memories should be represented, abstracted, and reused, instead of obtaining fixed or predefined abstract memory itself. As shown in Figure \ref{fig:mcma examples}, MCMA is built around a \textbf{\textit{Memory Copilot}} that operates at a meta level, learning to regulate the structure and granularity of memory and managing past successes and failures into reusable abstract memories. To enable this abstraction process to be learned and transferred independently of task-specific behaviors, MCMA decouples memory management from task execution by employing a frozen \textbf{\textit{Task Model}} solely for action selection.

MCMA follows a four-stage pipeline: collecting trajectories, generating structured memories, training abstraction strategies, and reusing knowledge and ability. The memory copilot is trained via direct preference optimization (DPO) \cite{rafailov2023direct} to learn \textbf{multi-structural memory representations} (e.g., tree, chain, or natural language structures) and, crucially, \textbf{abstraction strategies that support memory reuse}. 
Guided by cognitive theories of memory abstraction, memory is commonly divided into \textit{episodic memory}, which stores concrete, context-dependent experiences, and \textit{semantic memory}, which encodes abstract, organized knowledge \cite{tulving1972episodic}. 
MCMA organizes structured memories into \textbf{multiple abstraction levels}, where lower levels retain fine-grained execution details, while higher levels save abstracted knowledge. 
When no prior memory is reusable, \textbf{the memory copilot itself is transferred}, preserving the learned abstraction capability. Experiments on ALFWorld ($\uparrow25.07\%$), ScienceWorld ($\uparrow7.92\%$), and BabyAI demonstrate substantial gains in robustness, out-of-distribution generalization, and cross-task transfer.

In summary, our contributions are as follows:
\begin{itemize}
    \item This paper proposes a meta-cognitive memory abstraction method \textbf{MCMA}. By learning structural abstract memory representations and organizing reusable hierarchical abstractions, MCMA transforms memories from inflexible storage into a transferable resource.
    \item MCMA introduces memory copilot, trained via DPO to distill trajectories into structured abstract knowledge that is more suitable for assisting unseen tasks. Crucially, the memory copilot itself can be transferred across domains, reusing the learned ability to abstract, reflect, and organize new tasks.
    \item Extensive experiments on ALFWorld, ScienceWorld, and BabyAI, demonstrating substantial improvements in basic performance, out-of-distribution generalization, and cross-task transfer.
\end{itemize}

\begin{figure*}[t]
\centering
  \includegraphics[width=1\linewidth]{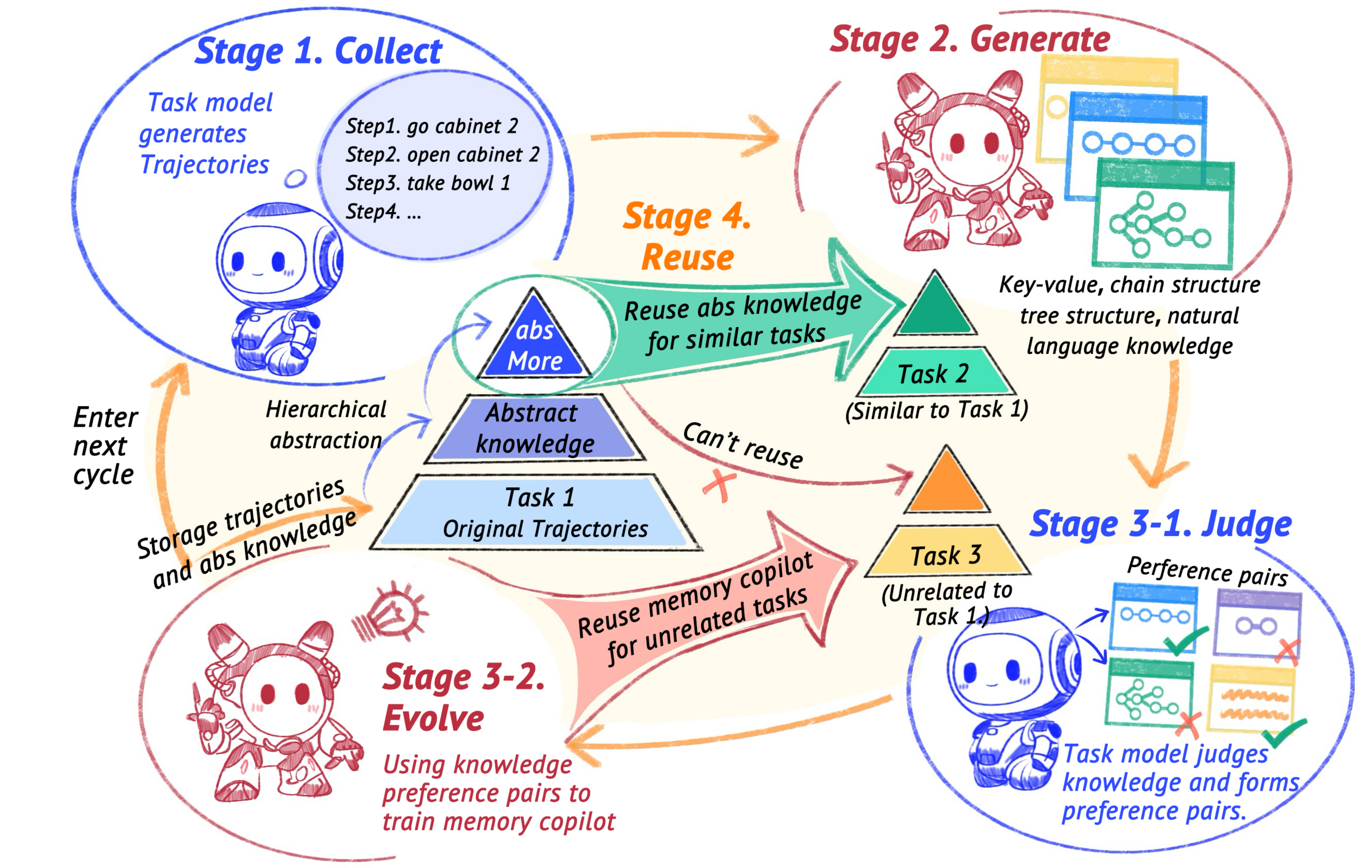}
  \caption {The task model collects raw trajectories, which the memory copilot abstracts into structured knowledge. Preference pairs constructed from downstream task performance are used to train the memory copilot via DPO. Learned memories and the copilot are organized hierarchically to support adaptive memory reuse across tasks.}
  \label{fig_main}
  \vspace{-8pt}
\end{figure*}

\section{Related Work}

\paragraph{Memory as Content Retrieval or Execution Guidance.} Prior work on memory-augmented LLM agents primarily focuses on storing and retrieving past experience to guide future decision making. Retrieval-based methods, including trajectory-level retrieval \cite{zheng2023synapse, kong2025mapagent, ritchford2025optimizing, luo2025agentauditor}, subtask or case-based reuse \cite{kim2024rada}, and step-wise retrieval aligned with intermediate reasoning states \cite{zhou2024trad}. While related efforts extract procedural patterns or heuristics from execution histories for planning. These approaches treat memory as reusable content with fixed representations and retrieval granularities. Consequently, retrieved experience often overfits to surface-level similarity and can mislead agents under changes in environment or task, leading to a negative impact.

\paragraph{Hierarchical, Reflective, and Learned Experience}
Beyond direct retrieval, several works explore hierarchical organization, reflection, or learning from experience to improve agent robustness. Hierarchical and control-based approaches introduce multiple memory levels or regulate memory access to separate high-level planning from low-level execution \cite{ye2025h, peiyuan2024agile, zhao2024expel, wang2024agent, zhang2025g, xu2025kodcode}, while reflection-based methods refine future behavior through error correction \cite{song2024trial, wang2025steca, fu2024autoguide, shinn2023reflexion}. Large-scale training efforts further internalize experience by adjusting model parameters to improve generalization \cite{song2024agentbank, zeng2024agenttuning, wang2025cogito, yao2023retroformer, yin2024agent}. Despite these advances, memory representations and abstraction levels remain largely predefined or implicitly fixed, leaving agents without the ability to learn how experience should be abstracted or which abstraction level is fit for a new task. 

\section{Methodology}

As shown in Figure \ref{fig_main}, MCMA consists of two functionally distinct models: a \textbf{\textit{Task Model}} and a \textbf{\textit{Memory Copilot}}. 
\textit{Stage 1}, the task model collects interaction trajectories by performing the task. \textit{Stage 2}, the memory copilot transforms raw interaction trajectories into structured and reusable memories through selective structure combination and abstraction. \textit{Stage 3}, the task model evaluates these memories based on downstream task performance, constructs preference pairs to train the memory copilot, enabling its continual evolution. \textit{Stage 4}, MCMA organizes the accumulated memories into a hierarchy of abstractions and supports the reuse of either stored knowledge or the memory copilot itself.
For example, in ALFWorld, once a subtask is completed, the memory copilot abstracts the trajectory, and when the agent later encounters a similar task, an appropriate level of abstraction is selected for retrieval. 

\subsection{Trajectory Collection and Preprocessing}

The task model collects both successful and failed trajectories from training environments. Raw trajectories are typically long, noisy, and dominated by low-level execution details, making them inefficient for direct reuse. Given a trajectory, execution traces are converted into goal oriented representations via trajectory simplification: \textbf{1) Noise pruning}: remove redundant or task-irrelevant steps. \textbf{2) Subtask segmentation}: partition the remaining sequence into coherent subtask units based on changes in the goal (\textit{e.g.}, task \textit{"Put a bowl in the dining table"} can be divided into subtasks: \textit{"Take a bowl from the cabinet"}, and \textit{"Place a bowl on the dining table"}.). \textbf{3) Consistency checking}: Verify logical dependencies and temporal ordering between subtasks. \textbf{4) Structured processing}: Organize subtasks into a tree, with high-level goals as parent nodes and executable skills as leaves. This process preserves both high-level intent and low-level executability while significantly reducing representation complexity. Appendix \ref{appendix_process} provides a detailed preprocessing procedure and examples.

\subsection{Multi-Structure Memory Generation}

The memory copilot learns a \textbf{meta-level abstraction policy} that jointly determines the \emph{structural composition} and \emph{level} of memory representations. Since memory representation critically affects reuse and transfer across tasks, we avoid assuming a fixed memory storage structure. Instead, for each trajectory $\tau$, the copilot generates a set of candidate \emph{composite structured memories}:
\begin{equation}
\mathcal{M}_\tau = \{ m_\tau^1, m_\tau^2, \dots, m_\tau^N \}.
\end{equation}
Each candidate $m_\tau^i$ is constructed by composing and nesting multiple structural primitives drawn from a predefined set:
\begin{equation}
\mathcal{S} = \{\text{natural text, key--value, chain, tree}\},
\end{equation}
rather than selecting a single structure type. Appendix \ref{appendix_structure_knowledge} provides several structural knowledge examples. Intuitively, different structure capture complementary aspects of experience: tree and chain structures encode high-level dependencies, while key-value pairs and natural language retain fine-grained details. Their composition enables expressive yet reusable memory representations. Each composite structured memory is evaluated by its downstream utility $s(m_\tau^i)$ (Equation \ref{equ_abs}), and the probability of generating a particular composition is defined as:
\begin{equation}
P_\theta(m_\tau^i \mid \tau) = \frac{\exp(\beta \, s(m_\tau^i))}{\sum_{j=1}^N \exp(\beta \, s(m_\tau^j))},
\end{equation}
where $\theta$ denotes the copilot parameters and $\beta > 0$ is a temperature parameter. This distribution is used to optimize memory selection toward representations with higher expected downstream utility.

To control abstraction granularity, we introduce a continuous abstraction parameter $\alpha \in [0,1]$:
\begin{equation}
m_\tau(\alpha) = \text{\textit{Memory Copilot }}_\theta^{\text{abs}}(\tau, \alpha).
\end{equation}
Smaller $\alpha$ preserves execution details, while larger $\alpha$ promotes higher procedure level abstractions. The memory copilot should learn an optimal $\alpha$, which maximizes downstream utility via training:
\begin{equation}
\alpha^* = \arg\max_{\alpha \in [0,1]} s\big(m_\tau(\alpha)\big).
\end{equation}
We collect a diverse set of multi-structure memories with varying abstraction levels to support subsequent memory copilot evolution.

\subsection{Memory Copilot Evolution}

For each trajectory, multiple knowledge candidates are judged on downstream tasks, producing scores:
\begin{equation}
\label{equ_abs}
s(m_i) =
\begin{cases}
0, & \text{task fails}, \\
0.1 + 0.9 \cdot \dfrac{T_{\max} - T_i}{T_{\max} - T_{\min}}, & \text{otherwise},
\end{cases}
\end{equation}
where $T_i$ denotes execution length. The $Top\text{-}K$ preference pairs with the largest differences in $s(m_i)$ are used for training. This strategy introduces strong supervision signals while preserving diversity among memory representations. We adopt a two-stage procedure consisting of supervised fine-tuning followed by direct preference optimization (DPO) \cite{rafailov2023direct}. For a preference pair $(m^+, m^-)$ derived from the same trajectory $\tau$, the DPO objective is:
\begin{equation}
\begin{aligned}
\mathcal{L} =
& - \log \sigma \Big(
\beta \big[
\log p_\theta(m^+) - \log p_\theta(m^-)
\big]
\Big)
\end{aligned}
\end{equation}
where $p_\theta(m)$ denotes the likelihood assigned by the memory copilot parameterized by $\theta$ to generating a structured memory abstraction $m$ conditioned on trajectory $\tau$. Minimizing this objective encourages the model to assign higher probability to memory representations that lead to more successful and efficient downstream task execution. Success and failure trajectories are processed separately, resulting in two memory copilots with different functions (successful memory summarization and failure memory reflection). During inference, $Top\text{-}N$ relevant trajectories are retrieved using character-level matching on task descriptions, then abstracted by the copilot before being provided to the task model.

\subsection{Hierarchical Abstraction and Cross-Task Reusing}

After learning instance-level memory representations, memories are organized into a hierarchy:
\begin{equation}
\mathcal{H} = \{H_0, H_1, \dots, H_L\},
\end{equation}
where lower levels store detailed memories, $H_0$ retaining raw trajectories, low levels (\textit{e.g.}, $H_1$) encoding episodic memories with execution details, and higher levels (\textit{e.g.}, $H_L$) capturing abstract knowledge such as scripts ($L=2$ in our work). We perform similarity clustering based on the vector representation of the task description. High-level abstractions are obtained by merging lower-level representations while suppressing task-specific details to retain shared goals. For a new task $\tau_{\text{new}}$, memory reuse is performed via:
\begin{equation}
m_{\text{reuse}} = \arg\max_{m \in H_\ell, \ell \in [0,L]} \text{sim}(\tau_{\text{new}}, m).
\end{equation}
High-similarity tasks are supported by low-level, detailed episodic memories, whereas low-similarity tasks use higher-level abstract memories. Abstracted knowledge is incorporated into the prompt to assist the model when solving similar tasks.

Crucially, under extreme distribution shifts where no stored memory is directly reusable, MCMA transfers the memory copilot itself rather than specific trajectories. Let $\mathcal{T}_{\text{train}}$ and $\mathcal{T}_{\text{new}}$ denote the training and novel task distributions. The memory copilot learns a transferable abstraction strategy $\theta^*$ by optimizing
\begin{equation}
\theta^* = \arg\max_\theta \mathbb{E}_{\tau \sim \mathcal{T}_{\text{train}}}
\big[ s(\text{Abstraction}_\theta(\tau, \alpha)) \big].
\end{equation}
By training, $\theta^*$ can be directly applied to new tasks $\tau_{\text{new}} \sim \mathcal{T}_{\text{new}}$ to generate abstract memories, enabling the model to generalize and transfer effectively even in the absence of relevant stored trajectories.

In this way, the memory copilot learns not only what to remember but how experience should be abstracted to maximize its future utility, enabling robust transfer across tasks and domains.

\section{Experiment}
 
\newcommand{\upgood}[1]{$_{\color{mygreen}\uparrow #1}$}
\newcommand{\downbad}[1]{$_{\color{myred}\downarrow #1}$}
\newcommand{\downgood}[1]{$_{\color{mygreen}\downarrow #1}$}
\newcommand{\upbad}[1]{$_{\color{myred}\uparrow #1}$}

\begin{table*}[t]
\centering
\setlength{\tabcolsep}{2.5pt}
\resizebox{\linewidth}{!}{
\begin{tabular}{ccllllllll}
\toprule
\multirow{3}{*}{\textbf{Model}} & \multirow{3}{*}{\textbf{Method}} & \multicolumn{4}{c}{\textbf{ALFWorld}} & \multicolumn{2}{c}{\textbf{Science World}} \\
\cmidrule(lr){3-6} \cmidrule(lr){7-8}
 & & \multicolumn{2}{c}{\textbf{Seen}} & \multicolumn{2}{c}{\textbf{Unseen}} & \multicolumn{1}{c}{\textbf{Dev}} & \multicolumn{1}{c}{\textbf{Test}} \\
\cmidrule(lr){3-4} \cmidrule(lr){5-6} \cmidrule(lr){7-8} 
 & & \textbf{Acc}(\%) $\uparrow$ & \textbf{Step} $\downarrow$ & \textbf{Acc}(\%) $\uparrow$ & \textbf{Step} $\downarrow$ & \textbf{Reward} $\uparrow$ & \textbf{Reward} $\uparrow$ \\
\midrule
\multirow{6}{*}{\textbf{Qwen3-8B}} 
 & \textbf{No Memory}             & 55.00 & 33.15 & 56.33 & 33.06 & 21.26 & 19.18 \\
 & \textbf{ReAct}                 & 59.29 \upgood{4.29} & 31.04 \downgood{2.11} & 64.93 \upgood{8.60} & 30.58 \downgood{2.48} & 21.00 \downbad{0.26} & 18.41 \downbad{0.77} \\
 & \textbf{Raw Tra}     & 67.14 \upgood{12.14} & 24.75 \downgood{8.40} & \cellcolor{secondcolor}72.39 \upgood{16.06} & 25.01 \downgood{8.05} & 20.17 \downbad{1.09} & 17.61 \downbad{1.57} \\
 & \textbf{TRAD}                  & 64.29 \upgood{9.29} & 28.48 \downgood{4.67} & 63.43 \upgood{7.10} & 29.76 \downgood{3.30} & \cellcolor{secondcolor}30.16 \upgood{8.90} & \cellcolor{secondcolor}27.42 \upgood{8.24} \\
 & \textbf{ExpeL}                 & \cellcolor{secondcolor}69.28 \upgood{14.28} & \cellcolor{secondcolor}24.73 \downgood{8.42} & \cellcolor{secondcolor}72.39 \upgood{16.06} & \cellcolor{secondcolor}24.16 \downgood{8.90} & 26.43 \upgood{5.17} & 23.12 \upgood{3.94} \\
 \cmidrule(lr){2-8} 
 & \textbf{MCMA}$_{\textit{(ours)}}$       & \cellcolor{bestcolor} \textbf{79.29} \upgood{24.29} & \cellcolor{bestcolor} \textbf{21.24} \downgood{11.91} & \cellcolor{bestcolor} \textbf{80.60} \upgood{24.27} & \cellcolor{bestcolor} \textbf{20.57} \downgood{12.49} & \cellcolor{bestcolor} \textbf{31.17} \upgood{9.91} & \cellcolor{bestcolor} \textbf{29.17} \upgood{9.99} \\

\midrule
\multirow{6}{*}{\textbf{Qwen3-32B}} 
 & \textbf{No Memory}             & 62.86 & 29.85 & 66.42 & 30.36 & 45.10 & 43.69 \\
 & \textbf{ReAct}                 & 63.57 \upgood{0.71} & 31.15 \upbad{1.30} & 72.39 \upgood{5.97} & 28.75 \downgood{1.61} & \cellcolor{secondcolor}45.44 \upgood{0.34} & \cellcolor{secondcolor}44.05 \upgood{0.36} \\
 & \textbf{Raw Tra}     & 74.29 \upgood{11.43} & 21.92 \downgood{7.93} & 78.36 \upgood{11.94} & 21.39 \downgood{8.97} & 44.82 \downbad{0.28} & 41.70 \downbad{1.99} \\
 & \textbf{TRAD}                  & 74.29 \upgood{11.43} & 24.77 \downgood{5.08} & \cellcolor{secondcolor}79.10 \upgood{12.68} & 24.73 \downgood{5.63} & 42.77 \downbad{2.33} & 42.00 \downbad{1.69} \\
 & \textbf{ExpeL}                 & \cellcolor{secondcolor}78.57 \upgood{15.71} & \cellcolor{secondcolor}20.40 \downgood{9.45} & 77.61 \upgood{11.19} & \cellcolor{secondcolor}21.23 \downgood{9.13} & 44.97 \downbad{0.13} & 43.78 \upgood{0.09} \\
 \cmidrule(lr){2-8} 
 & \textbf{MCMA}$_{\textit{(ours)}}$ & \cellcolor{bestcolor} \textbf{90.71} \upgood{27.85} & \cellcolor{bestcolor} \textbf{17.78} \downgood{12.07} & \cellcolor{bestcolor} \textbf{90.30} \upgood{23.88} & \cellcolor{bestcolor} \textbf{18.00} \downgood{12.36} & \cellcolor{bestcolor} \textbf{51.95} \upgood{6.85} & \cellcolor{bestcolor} \textbf{48.60} \upgood{4.91} \\
\bottomrule
\end{tabular}
}
\caption{Performance comparison of the Qwen3 model. We highlight the \colorbox{bestcolor}{best} and \colorbox{secondcolor}{second best} results. Green arrows (${\color{mygreen}\uparrow}/{\color{mygreen}\downarrow}$) indicate performance improvement, and Red arrows (${\color{myred}\uparrow}/{\color{myred}\downarrow}$) indicate decline.}
\label{tab:main_results}
\end{table*}

\begin{table}[t]
    \centering
    \begin{tabular}{cll}
        \toprule
        \textbf{Model} & \textbf{Seen} & \textbf{Unseen} \\
        \midrule
        \textit{GPT-4o-mini} & 42.86 & 48.51 \\
        \textbf{MCMA}$_{\textit{(GPT)}}$ & \textbf{63.57} \upgreen{20.71} & \textbf{63.43} \upgreen{14.92} \\
        \midrule
        \textit{Gemini-2.5-flash} & 61.43 & 67.91 \\
        \textbf{MCMA}$_{\textit{(Gemini)}}$ & \textbf{84.29} \upgreen{22.86} & \textbf{88.06} \upgreen{20.15} \\
        \bottomrule
    \end{tabular}
    \caption{Reuse memory copilot of Qwen3-32B on \textit{Gemini-2.5-flash} and \textit{GPT-4o-mini} on ALFWorld task.}
    \vspace{-10pt}
    \label{tab:gemini}
\end{table}

\subsection{Experimental Setup}

\paragraph{Datasets.}
We evaluate MCMA on two long-horizon text-based embodied reasoning benchmarks: ALFWorld \cite{shridhar2020alfworld} and ScienceWorld \cite{wang2022scienceworld}. ALFWorld evaluates household task completion from textual observations and provides \textit{seen} and \textit{unseen} splits. The \textit{seen} split contains the same task types, objects, and room categories as training but varies object configurations. The \textit{unseen} task instances are executed in rooms that never appear during training. ScienceWorld focuses on multi-step scientific reasoning under complex text-based dynamics and follows a standard \textit{Dev / Test} split. Both benchmarks emphasize long-horizon reasoning and generalization beyond memorizing surface-level execution patterns, requiring effective abstraction, experience reuse, and adaptation to unseen environments.

\paragraph{Metrics.}
For ALFWorld, we follow the standard evaluation method and report results on both seen and unseen task splits. Performance is measured by task success rate (Acc\%) and average execution steps, where higher accuracy and fewer steps indicate better performance. For ScienceWorld, we evaluate on the Dev and Test splits and report the average task reward score, which reflects the progress of sub-goals.

\paragraph{Models.}
We conduct experiments using two backbone task models: Qwen3-8B \cite{yang2025qwen3} and Qwen3-32B. The memory copilot uniformly uses Qwen3-4B. In all settings, the task model is frozen during training and evaluation, and all memory-related learning is isolated within the memory copilot. 

\paragraph{Baselines.}
We compare MCMA against several memory configurations: \textit{1) No Memory}, where the agent operates purely reactively without access to prior experience. \textit{ 2) ReAct} \cite{yao2022react}, adopt a process of fully observing the environment and thinking before making a decision. \textit{3) Raw Trajectory}, which retrieves and injects raw trajectories into memory. \textit{4) TRAD} \cite{zhou2024trad}, retrieves relevant expert steps via thought matching and aligning them with localized temporal context for each steps. \textit{5) ExpeL}  \cite{zhao2024expel}, which extracts high-level natural language insights from past experiences and leverages retrieved successful trajectories as in-context demonstrations during inference. For the ALFWorld task, MCMA uses both the successful memory summary and the failed memory reflection memory copilot. Considering that the ScienceWorld environment is more variable, and summarizing successful memories often has negative effects, this task only used the failed memory reflection memory copilot.

\subsection{Main Results}

As shown in Table~\ref{tab:main_results}, MCMA consistently achieves the strongest performance across all environments and model scales. Compared to all baselines, MCMA significantly improves task success rates while reducing execution steps, indicating more efficient and reliable long-horizon planning.

\textbf{Consistent gains across seen and unseen tasks.}
MCMA delivers large and stable improvements over no-memory and prior memory-based baselines in both in-domain and out-of-distribution settings. For example, on ALFWorld with Qwen3-8B, MCMA improves success rates by nearly 25\% over the no-memory baseline, with comparable gains on unseen tasks. MCMA maintains similar improvements across seen and unseen scenarios, highlighting the strong generalization of its meta-cognitive memory abstraction.

\textbf{Improved efficiency and scalability across models.}
MCMA consistently reduces the average number of execution steps across benchmarks, producing efficient and concise action sequences. These benefits are particularly pronounced for smaller models: on ScienceWorld, MCMA yields nearly a 10\% absolute gain on Qwen3-8B, while still providing consistent improvements on Qwen3-32B. This indicates that memory abstraction effectively assists capacity limited model and remains powerful even as model scale increases. We further test the performance of the memory copilot trained for Qwen3-32B on the closed-source model Gemini-2.5-Flash and GPT-4o-mini. As shown in Table \ref{tab:gemini}, the performance improvement is close to that achieved on Qwen3-32B, indicating that MCMA has good transferability and performs well on closed-source models.

\subsection{Ablation Study}

\begin{table*}[htbp]
    \centering
    \begin{tabular}{clllll}
        \toprule
        \multirow{3}{*}{\textbf{Model}} & \multirow{3}{*}{\textbf{Configuration}} & \multicolumn{2}{c}{\textbf{Seen}} & \multicolumn{2}{c}{\textbf{Unseen}} \\
        \cmidrule(lr){3-4} \cmidrule(lr){5-6}
        & & \textbf{Acc(\%)} $\uparrow$ & \textbf{Step} $\downarrow$ & \textbf{Acc(\%)} $\uparrow$ & \textbf{Step} $\downarrow$ \\
        \midrule
        \multirow{5}{*}{\textbf{Qwen3-8B}} & \textbf{Natural Language Know} & 74.63 \downred{4.66} & 23.37 \upred{2.13} & 72.39 \downred{8.21} & 24.88 \upred{4.31} \\
        & \textbf{Chain Know} & 77.14 \downred{2.15} & 22.57 \upred{1.33} & 75.27 \downred{5.23} & 23.55 \upred{2.98} \\
        \cmidrule(lr){2-6}
        & \textbf{MCMA }$_{\textit{(Sum)}}$ & 68.57 \downred{10.72} & 23.69 \upred{2.45} & 72.39 \downred{8.21} & 23.85 \upred{3.28} \\
        & \textbf{MCMA }$_{\textit{(Ref)}}$ & 72.86 \downred{6.43} & 25.15 \upred{3.91} & 71.64 \downred{8.96} & 25.82 \upred{5.25} \\
        & \textbf{MCMA }$_{\textit{(Sum+Ref)}}$ & \textbf{79.29} & \textbf{21.24} & \textbf{80.60} & \textbf{20.57} \\
        \midrule
        \multirow{5}{*}{\textbf{Qwen3-32B}} & \textbf{Natural Language Know} & 77.86 \downred{12.85} & 20.50 \upred{2.72} & 82.09 \downred{8.21} & 21.75 \upred{3.75} \\
        & \textbf{Chain Know} & 86.14 \downred{4.57} & 19.10 \upred{1.32} & 84.33 \downred{5.97} & 20.24 \upred{2.24} \\
        \cmidrule(lr){2-6}
        & \textbf{MCMA }$_{\textit{(Sum)}}$ & 76.43 \downred{14.28} & 21.84 \upred{4.06} & 83.25 \downred{7.05} & 23.97 \upred{5.97} \\
        & \textbf{MCMA }$_{\textit{(Ref)}}$ & 82.14 \downred{8.57} & 22.46 \upred{4.68} & 84.33 \downred{5.97} & 21.95 \upred{3.95} \\
        & \textbf{MCMA }$_{\textit{(Sum+Ref)}}$ & \textbf{90.71} & \textbf{17.78} & \textbf{90.30} & \textbf{18.00} \\
        \bottomrule
    \end{tabular}
    \caption{Component and Memory Representation ablation study of Qwen3-32B settings on ALFWorld .}
    \label{tab:Ablation}
\end{table*}

\paragraph{Component Ablations.}
To analyze the contribution of individual components in the memory copilot, we conduct ablation studies on ALFWorld, focusing on successful memory summarization (\textit{Sum}) and failure memory reflection (\textit{Ref}). As shown in Table~\ref{tab:Ablation}, each component independently improves performance: summarization reduces execution steps by abstracting procedural knowledge, while reflection helps avoid recurring errors. Combining both components achieves the best performance across all settings and model scales, yielding higher success rates with fewer steps, particularly under distribution shift. These results suggest that summarization and reflection capture complementary aspects of experience \textit{"learning what to do and what to avoid"}, and are jointly critical for robust generalization.

\begin{table}[t]
    \centering
    \begin{tabular}{lll}
        \toprule
        \textbf{Variant} & \textbf{Acc(\%)} $\uparrow$ & \textbf{Step} $\downarrow$ \\
        \midrule
        \textbf{MCMA}$_{\textit{(Base)}}$ & 83.58 \downred{7.13}  & 22.53 \upred{4.75} \\
        \textbf{MCMA}$_{\textit{(Gemini)}}$ & 86.56 \downred{4.15}  & 19.75 \upred{1.97} \\
        \midrule
        \textbf{MCMA}$_{\textit{(DPO)}}$  & \textbf{90.71}  & \textbf{17.78}     \\
        \bottomrule
    \end{tabular}
    \caption{Comparison with MCMA using untrained Qwen3-4B / Gemini-2.5-flash as memory copilot.}
    \label{tab:architecture ablation}
    \vspace{-10pt}
\end{table}

\paragraph{Memory Representation and Training Ablations.}
We further examine the impact of memory representation and training by comparing MCMA with variants that store experiences as natural language or chain-structured knowledge generated by DPO-trained memory copilots. As shown in Table~\ref{tab:Ablation} (\textit{Nautral Language Know} and \textit{Chain Know}), both representations yield substantial gains over the no-memory baseline (average accuracy 64.64\%), suggesting that the proposed abstraction strategy is relatively insensitive to specific surface structures. Nevertheless, a consistent performance gap remains compared to the full MCMA (MCMA$_{\textit{(Sum+Ref)}}$), highlighting the importance of structured organization for effective memory utilization. In addition, when deploying untrained memory copilots Qwen3-4B or more powerful \textit{Gemini-2.5-flash} in unseen settings, performance drops 7.13\% and 4.15\% (Table~\ref{tab:architecture ablation}), indicating that DPO training is essential for equipping the memory copilot with transferable abstraction and generalization capabilities.

\subsection{Structured Memory Analyze}

\begin{figure}[t]
  \includegraphics[width=1\columnwidth]{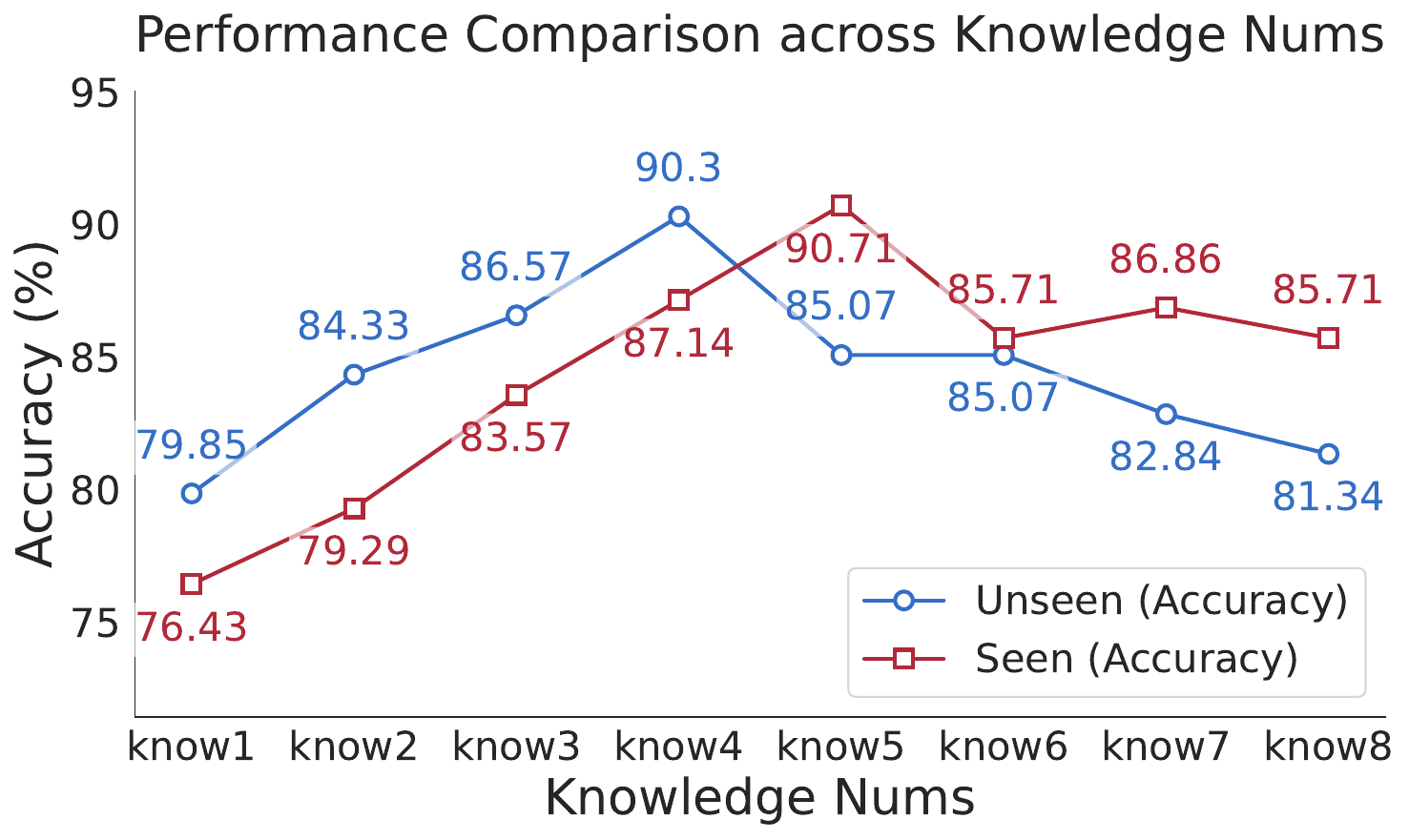}
  \caption{The impact of the number of knowledge provided by MCMA in the ALFWorld task.}
  \label{fig:knowledge num}
  \vspace{-10pt}
\end{figure}

The quantity of provided knowledge significantly impacts the model's performance. As illustrated in Figure \ref{fig:knowledge num}, MCMA achieves optimal results on ALFWorld when 4–5 structural knowledge are provided. Insufficient knowledge fails to equip the task model with enough foresight to adapt diverse scenarios, while excessive knowledge imposes a heavy contextual overhead that disturbs decision-making. Notably, MCMA’s structured knowledge is significantly more concise than original traces, it allows for a higher density of knowledge items within the prompt. 

\begin{figure}[t]
\centering
  \includegraphics[width=1\linewidth]{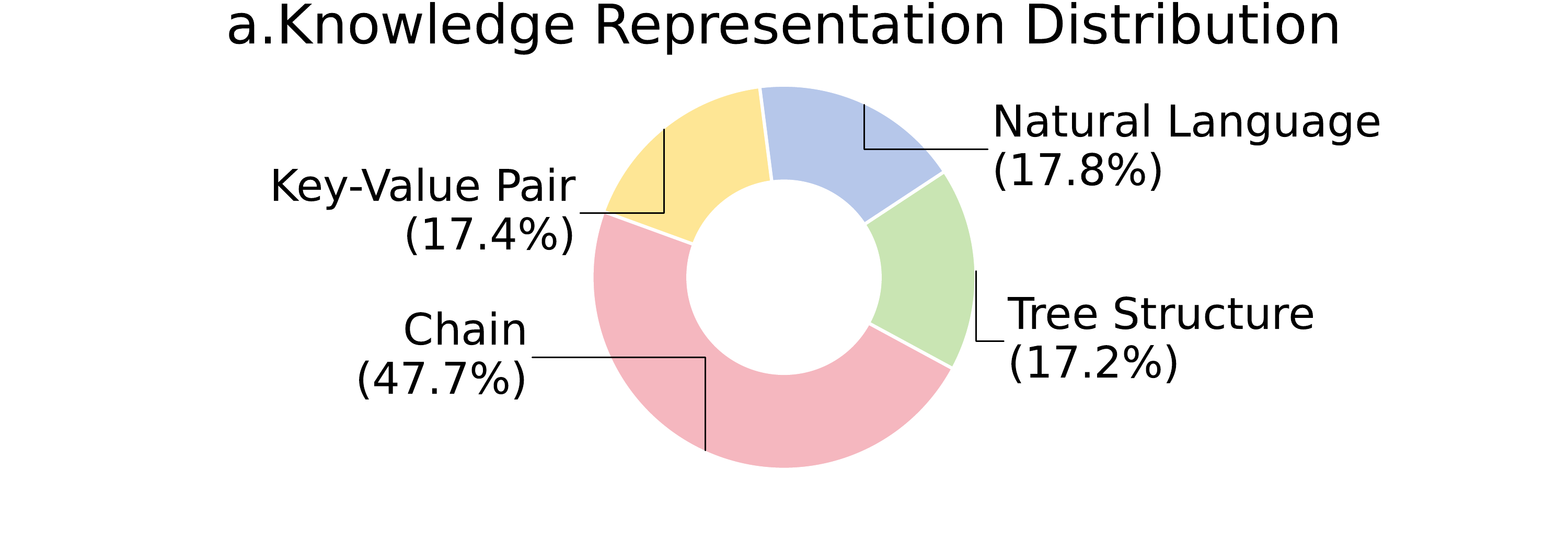}
  \vspace{-25pt}
  \includegraphics[width=0.95\linewidth]{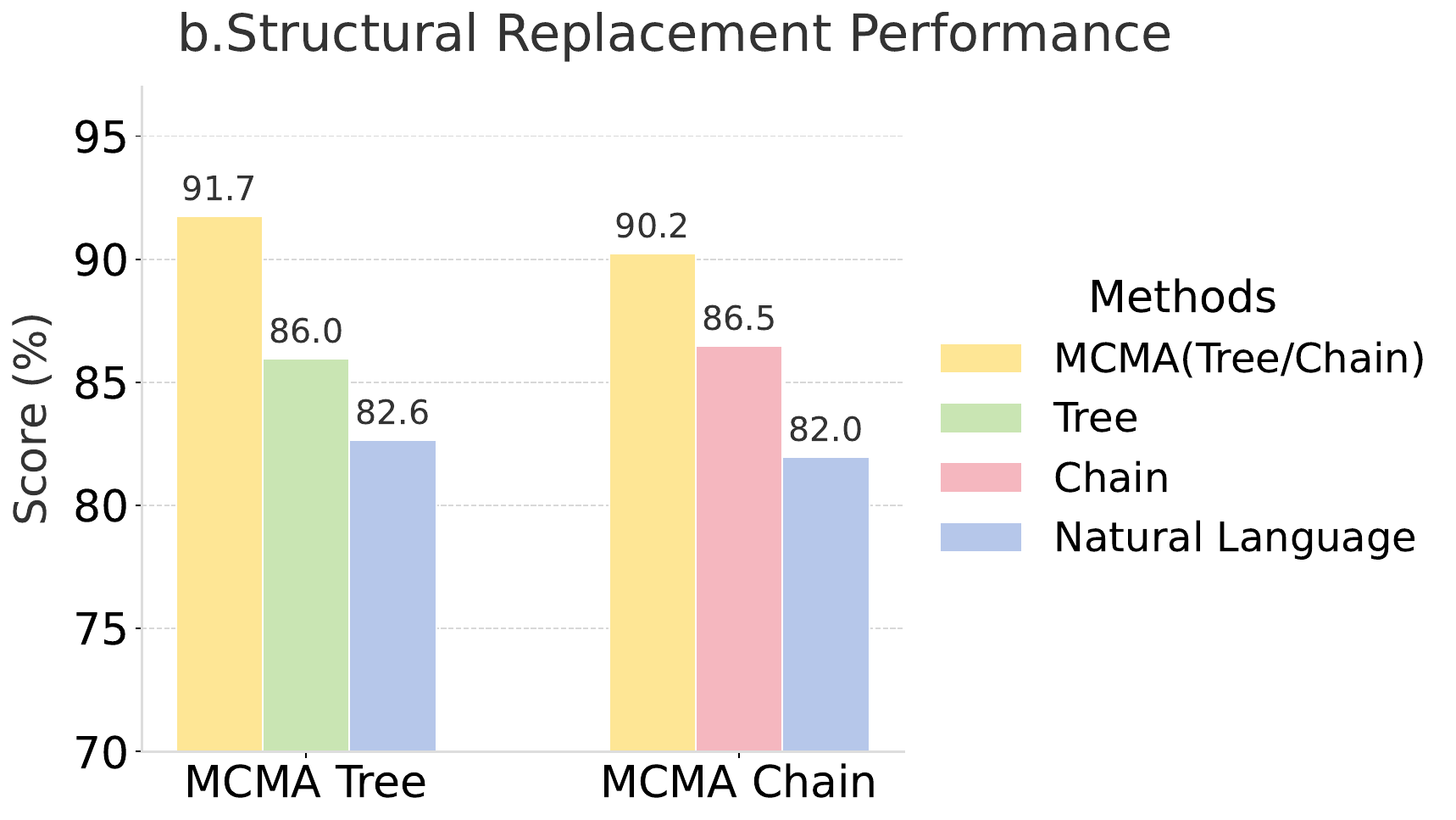}
  \vspace{20pt}
  \caption {Analyze knowledge structure of MCMA. }
  \label{fig:analyze}
  \vspace{-10pt}
\end{figure}

In addition, we count all nested structural knowledge. At the top-level, the predominant structures are Tree and Chain (Specifically, Tree structures comprised approximately 70\% of these top-level arrangements, while Chain structures accounted for about 20\%). This hierarchical organization effectively provides high-level directional guidance to the task model. Furthermore, within both Tree and Chain structures, a wide variety of sub-structures, such as Key-Value pairs and Natural Language descriptions, are nested, enabling a more fine-grained and explicit representation of detailed information. The overall distribution of all knowledge structure types is comprehensively illustrated in Figure \ref{fig:analyze}.a. We further identify tasks characterized by Tree and Chain top-level structure and replace the knowledge representations with other structural formats. As shown in Figure \ref{fig:analyze}.b, this led to a decline in performance, which underscores the effectiveness of the structure selection strategy studied by MCMA.

\subsection{Cross-Task Knowledge Transfer}

Previous sections have shown that MCMA generalizes effectively within the same task distribution. This section extends our investigation to evaluate how MCMA performs in entirely novel environments.

To evaluate the effectiveness of hierarchical knowledge transfer under distribution shift, we extend our evaluation to the BabyAI benchmark~\cite{chevalier2018babyai}. BabyAI requires agents to interpret and execute compositional, synthetic natural language instructions in a partially observable 2D gridworld, involving navigation, object manipulation, and environment interaction. Although BabyAI shares high-level similarities with ALFWorld in terms of spatial reasoning and object-centric interaction, it introduces a substantial domain shift, particularly in observation modalities (2D grid states vs.\ textual descriptions) and action granularity.

\begin{table}[htbp]
    \centering
    \begin{tabular}{cll} 
        \toprule
        \textbf{Variant} & \textbf{Acc(\%)} $\uparrow$ & \textbf{Reward} $\uparrow$ \\
        \midrule
        \textbf{Base}   & 16.67 & 14.24 \\
        \midrule
        \textbf{Abstract Level 1} & 13.54 \downred{3.13} & 12.32 \downred{1.92} \\
        \textbf{Abstract Level 2} & \textbf{17.71} \upgreen{1.04} & \textbf{15.33} \upgreen{1.09} \\
        \bottomrule
    \end{tabular}
    \caption{Apply ALFWorld knowledge on BabyAI.}
    \vspace{-5pt}
    \label{tab:babyai}
\end{table}

As shown in Table~\ref{tab:babyai}, transfer performance is strongly correlated with abstraction granularity. Fine-grained Level~1 knowledge leads to a 3.13\% performance drop, indicating negative transfer caused by mismatched low-level interaction details. In contrast, higher-level Level~2 abstractions improve the baseline by 1.04\%, suggesting that abstract knowledge better suppresses domain-specific noise and preserves reusable task semantics. These results demonstrate that higher-level abstractions are essential for effective generalization across distantly related tasks.

\subsection{Cross-Domain Copilot Transfer}

\begin{table}[htbp]
    \centering
    \resizebox{\linewidth}{!}{
        \begin{tabular}{cll}
            \toprule
            \textbf{Method} & \textbf{ScienceWorld} & \textbf{ALFWorld} \\
            \midrule
            \textbf{Base} & 19.18 & 56.33 \\
            \midrule
            \textbf{Sci Copilot} & 29.17 \upgreen{9.99} & 61.19 \upgreen{4.86} \\
            \textbf{ALF Copilot} & 27.43 \upgreen{8.25} & \textbf{71.64} \upgreen{15.31} \\
            \textbf{Mix Copilot} & \textbf{29.85} \upgreen{10.67} & 69.40 \upgreen{13.07} \\
            \bottomrule
        \end{tabular}
    }
    \caption{Copilots transfer across domains.}
    \vspace{-5pt}
    \label{tab:copilot-transfer}
\end{table}

We further explore a more challenging scenario: \textit{how to reuse memory when a new task has no correlation with existing memories}. MCMA achieves this by transferring the memory copilot, which reuses the capability to organize and abstract knowledge rather than the knowledge itself. We evaluate this via a cross-task transfer between ALFWorld and ScienceWorld. As shown in Table \ref{tab:copilot-transfer}, transferring copilots between ALFWorld and ScienceWorld consistently outperforms the baseline. ALFWorld copilot outperforms ScienceWorld copilot in overall performance, we speculate this is because ALFWorld collecting more training data. Notably, the Mix Copilot (trained on mix data collected from ALFWorld and ScienceWorld tasks) proves effective for both domains, particularly providing substantial gains for the task with limited training resources (ScienceWorld). These results provide compelling evidence that MCMA facilitates generalization by learning a transferable meta-cognitive skill for memory management.

\section{Conclusion}

We propose MCMA, a meta-cognitive memory abstraction method that treats experience reuse as a learnable problem. By decoupling task execution from memory management, MCMA uses a transferable memory copilot trained via preference optimization to abstract and organize interaction trajectories. Experiments on ALFWorld, ScienceWorld, and BabyAI show that MCMA improves task performance, generalization, and cross-task transfer, highlighting the importance of learning how to remember in LLM-based agents.

\section*{Limitations}
MCMA has achieved good performance and generalization, but it also comes with several limitations. First, MCMA introduces additional computational overhead during training, as constructing preference supervision requires generating and evaluating multiple candidate abstractions for each trajectory. While this cost is incurred offline and the learned memory copilot can be reused across tasks, the overall training pipeline remains more expensive than retrieval-based or single-representation memory methods. Second, although MCMA organizes experience into a hierarchy of abstraction levels to support flexible reuse, the current reuse process relies on a manually designed strategy to select which abstraction level should be applied to a new task. This limits full end-to-end adaptivity and suggests an important direction for future work: learning abstraction-level selection jointly with memory structuring and reuse.

\section*{Ethical consideration}
Our research explores the reuse of Large Language Models (LLMs) memory. Despite undergoing additional fine-tuning in various experiments, these models retain ethical and social risks inherent in their pretraining data. Notably, open-source LLMs may incorporate private or contentious data during the training phase, thereby raising additional ethical concerns.

\bibliography{custom}
\appendix

\section{Appendix}
\label{sec:appendix}

\subsection{Dataset Details}

\paragraph{ALFWorld.} \cite{shridhar2020alfworld}
We conduct experiments using the ALFWorld environment \cite{shridhar2020alfworld}, a framework designed to bridge abstract text-based reasoning and embodied execution. ALFWorld aligns the text-based TextWorld engine \cite{cote18textworld} with the visually grounded ALFRED benchmark \cite{ALFRED20} by utilizing the Planning Domain Definition Language (PDDL) to synchronize state dynamics across modalities. The dataset encompasses six types of compositional household tasks (e.g., \textit{Pick \& Place}, \textit{Clean \& Place}, \textit{Heat \& Place}) distributed across 120 interactive scenes, including kitchens, bedrooms, bathrooms, and living rooms. To evaluate generalization, ALFWorld defines strict data splits: a \textbf{Seen} set containing 140 task instances in rooms encountered during training, and an \textbf{Unseen} set containing 134 tasks in novel rooms with different layouts and object placements, thereby testing the agent's ability to perform zero-shot transfer in out-of-distribution environments.

\begin{figure}[h]
    \centering
    \begin{tcolorbox}[
        colback=white, 
        colframe=boxPurple, 
        title=\textbf{ALFWorld Task Example},
        fonttitle=\bfseries,
        boxrule=1pt
    ]
    \small
    \textbf{Task Description:} Examine the alarmclock with the desklamp. \\
    \textbf{Initial Observation:} You are in the middle of a room. Looking quickly around you, you see a bed 1, a desk 1, a drawer 3, \dots and a sidetable 2.
    
    \vspace{0.2cm}
    \hrule
    \vspace{0.2cm}
    
    \texttt{> go to desk 1} \\
    \textbf{Observation:} You arrive at desk 1. On the desk 1, you see a \textbf{alarmclock 1}, a bowl 3, a cellphone 2, \dots and a pencil 2.
    
    \vspace{0.15cm}
    \texttt{> take alarmclock 1 from desk 1} \\
    \textbf{Observation:} You pick up the alarmclock 1 from the desk 1.
    
    \vspace{0.15cm}
    \texttt{> go to sidetable 2} \\
    \textbf{Observation:} You arrive at sidetable 2. On the sidetable 2, you see a cellphone 1, a \textbf{desklamp 1}, and a keychain 2.
    
    \vspace{0.15cm}
    \texttt{> use desklamp 1} \\
    \textbf{Observation:} You turn on the desklamp 1. 
    
    \vspace{0.15cm}
    \textit{Goal Achieved.}
    \end{tcolorbox}
    \caption{A simplified trajectory of an agent performing an \textit{Examine in Light} task in ALFWorld.}
    \label{fig:alfworld_example}
\end{figure}

\paragraph{ScienceWorld.} \cite{wang2022scienceworld} ScienceWorld is a complex interactive text environment designed to evaluate the procedural reasoning capabilities of agents within the context of an elementary school science curriculum. Distinct from static question-answering benchmarks, ScienceWorld simulates a dynamic physical world with underlying engines for thermodynamics, electrical circuits, chemistry, and biological life cycles. The benchmark consists of 30 diverse tasks across 10 topics, ranging from testing electrical conductivity to conducting genetic experiments, requiring agents to synthesize declarative scientific knowledge into multi-step action sequences (e.g., \textit{heating a beaker}, \textit{connecting a battery}). There are 1819 tasks in the test set and 1796 tasks in the dev set. Furthermore, the environment enforces rigorous evaluation through thousands of parametric variations, testing an agent's ability to generalize scientific concepts to novel objects and scenarios.

\begin{figure}[t]
  \includegraphics[width=0.95\linewidth]{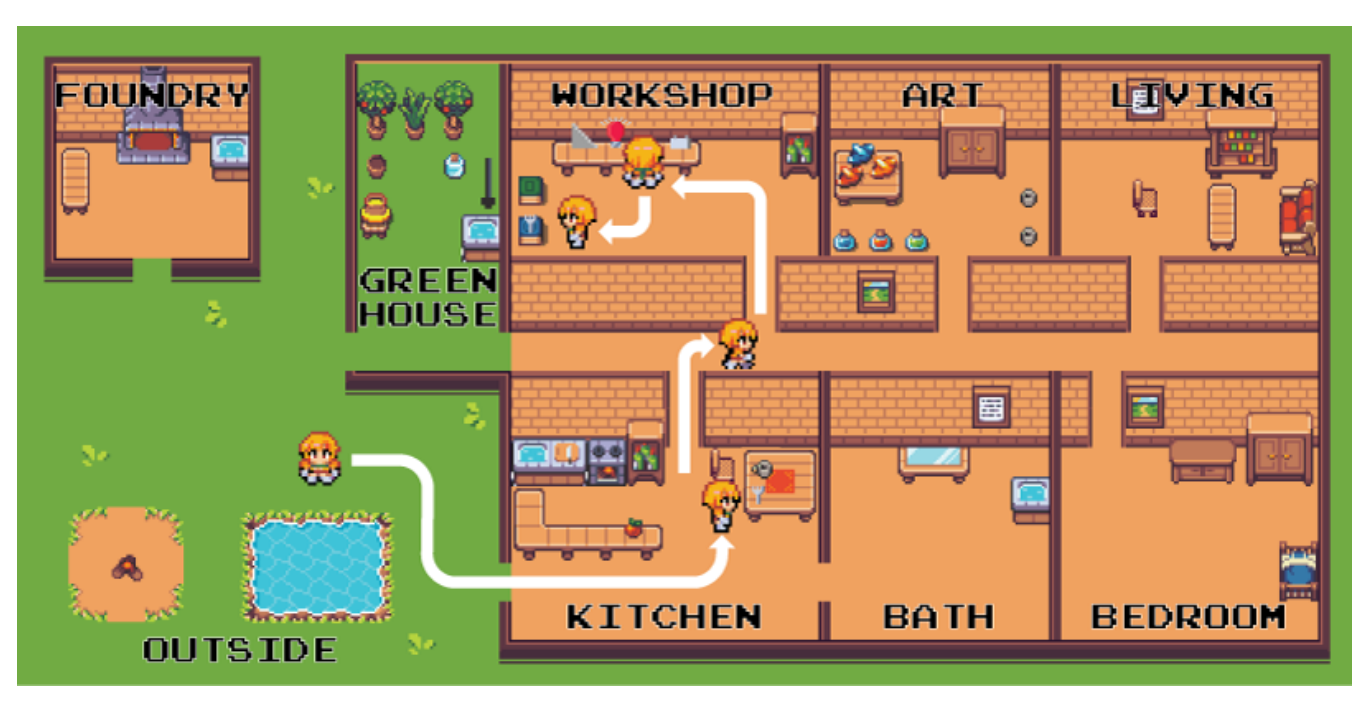}
  \caption {An example of ScienceWorld.}
  \label{fig_babyai}
\end{figure}

\begin{figure}[h]
    \centering
    \begin{tcolorbox}[
        colback=white, 
        colframe=boxPurple, 
        title=\textbf{ScienceWorld Interaction Example},
        fonttitle=\bfseries,
        boxrule=1pt
    ]
    \small
    \textbf{Task Description:} Find a living thing. First, focus on the thing. Then, move it to the orange box in the kitchen.
    
    \vspace{0.2cm}
    \hrule
    \vspace{0.2cm}
    
    \texttt{> look around} \hfill \textbf{[Score: 0.08]} \\
    \textbf{Observation:} This location is called the outside. You see a \textbf{blue jay egg}, a butterfly egg, a dove egg, an axe, and a door to the kitchen.
    
    \vspace{0.15cm}
    
    \texttt{> focus on blue jay} \hfill \textbf{[Score: 0.58]} \\
    \textbf{Observation:} You focus on the blue jay egg.
    
    \vspace{0.15cm}
    
    \texttt{> pick up blue jay} \hfill \textbf{[Score: 0.67]} \\
    \textbf{Observation:} You move the blue jay to the inventory.
    
    \vspace{0.15cm}
    
    \texttt{> go to kitchen} \hfill \textbf{[Score: 0.83]} \\
    \textbf{Observation:} You move to the kitchen. You see a counter, a table, and an \textbf{orange box}.
    
    \vspace{0.15cm}
    
    \texttt{> move blue jay to orange box} \hfill \textbf{[Score: 1.00]} \\
    \textbf{Observation:} You move the blue jay to the orange box.
    
    \vspace{0.15cm}
    \textit{Task Completed.}
    \end{tcolorbox}
    \caption{A step-by-step trajectory for the ``Find a living thing'' task in ScienceWorld. The score (right) updates incrementally as the agent completes sub-goals: identifying the correct object (focus), acquiring it, navigating, and placing it in the target container.}
    \label{fig:scienceworld_trace}
\end{figure}

\paragraph{BabyAI} \cite{chevalier2018babyai} We evaluate our transfer ability on the \textbf{BabyAI} platform \cite{chevalier2018babyai}, a benchmark designed to study the sample efficiency of grounded language learning agents. BabyAI is built upon the MiniGrid framework, featuring a partially observable 2D gridworld where agents perceive a $7 \times 7$ symbolic local view. The platform defines a curriculum of 19 levels of increasing difficulty, ranging from single-instruction tasks (e.g., \textit{GoToObj}) to complex composite missions involving navigation and manipulation (e.g., \textit{BossLevel}).

Figure \ref{fig_babyai} is an example of BabyAI's 2D grid image. To bridge the modality gap, we employ a deterministic translator that converts the agent's $7 \times 7$ egocentric symbolic observation into natural language. The agent is virtually anchored at the bottom-center coordinate $(3, 6)$ of the observable grid. For every salient object $i$ located at $(x_i, y_i)$ (excluding static background elements like walls), we calculate its relative spatial position using Manhattan distances: longitudinal offset $d_{\text{fwd}} = 6 - y_i$ and lateral offset $d_{\text{lat}} = x_i - 3$. These coordinates are mapped to egocentric linguistic templates (e.g., \textit{``A red ball is 3 steps forward and 1 step to the left''}), converting the sparse tensor into a dense descriptive list of visible entities. The final prompt fed to the LLM is constructed by concatenating the static system instructions, the specific \textit{Mission Goal}, a chronological \textit{History of Actions} $a_{0:t-1}$, and the \textit{Current Observation} generated above. The system instruction strictly defines the admissible action space (e.g., \textit{move forward}, \textit{toggle}) and enforces a structured JSON output format. This design ensures the model grounds its decision-making in both the immediate visual context and the temporal trajectory of the episode, allowing for chain-of-thought reasoning before action selection.

\begin{figure}[t]
  \includegraphics[width=0.95\linewidth]{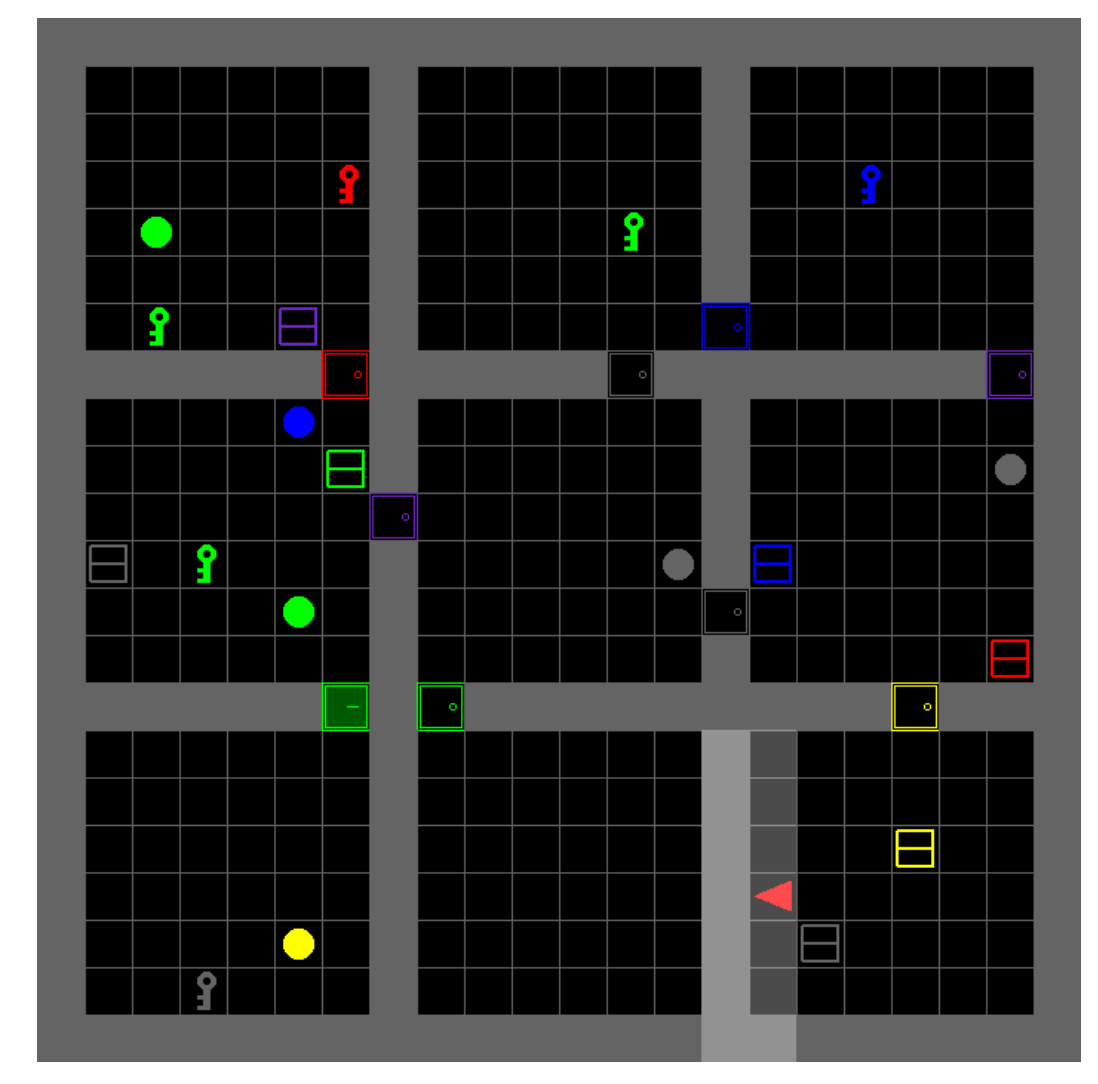}
  \caption {A 2D grid image example of BabyAI.}
  \label{fig_babyai}
\end{figure}

\begin{figure}[h]
    \centering
    \begin{tcolorbox}[
        colback=white, 
        colframe=boxPurple,
        title=\textbf{BabyAI Interaction Example},
        fonttitle=\bfseries,
        boxrule=1pt
    ]
    \small
    \textbf{Task Description:} Go to the green key.
    
    \vspace{0.2cm}
    \hrule
    \vspace{0.2cm}
    
    \texttt{> move forward} \hfill \textbf{[Score: 0.00]} \\
    \textbf{Observation:} You see a \textbf{green key} 3 steps forward and 2 steps to the left.
    
    \vspace{0.15cm}
    
    \texttt{> turn left} \hfill \textbf{[Score: 0.00]} \\
    \textbf{Observation:} You turn left. You see a \textbf{green key} 2 steps forward and 3 steps to the right.
    
    \vspace{0.15cm}
    
    \texttt{> move forward} \hfill \textbf{[Score: 0.00]} \\
    \textbf{Observation:} You see a \textbf{green key} 1 step forward and 3 steps to the right.
    
    \vspace{0.15cm}
    
    \texttt{> move forward} \hfill \textbf{[Score: 0.00]} \\
    \textbf{Observation:} You see a \textbf{green key} directly under you and 3 steps to the right.
    
    \vspace{0.15cm}
    
    \texttt{> turn right} \hfill \textbf{[Score: 0.00]} \\
    \textbf{Observation:} You turn right. You see a \textbf{green key} 3 steps forward.
    
    \vspace{0.15cm}
    
    \texttt{> move forward} \hfill \textbf{[Score: 0.00]} \\
    \textbf{Observation:} You see a \textbf{green key} 2 steps forward.
    
    \vspace{0.15cm}
    
    \texttt{> move forward} \hfill \textbf{[Score: 0.90]} \\
    \textbf{Observation:} You reach the goal.
    
    \vspace{0.15cm}
    \textit{Task Completed.}
    \end{tcolorbox}
    \caption{A step-by-step trajectory for the ``Go to the green key'' task in BabyAI. Unlike ScienceWorld, BabyAI typically uses sparse rewards, meaning the score (right) remains 0.00 until the specific goal condition is met at the final step.}
    \label{fig:babyai_trace}
\end{figure}

\subsection{Baseline Details}

\paragraph{ExpeL}
ExpeL \cite{zhao2024expel} is a non-parametric learning framework that enables LLM agents to improve autonomously through experience gathering and knowledge extraction. During a training phase, the agent collects successful and failed trajectories via trial-and-error (utilizing ReAct and Reflexion). It then abstracts cross-task knowledge into natural language \textit{insights} (e.g., guidelines or constraints) and stores successful trajectories in a vector database. At inference time, ExpeL augments the agent's context with these extracted insights and dynamically retrieves the top-$k$ most similar successful past trajectories to serve as few-shot examples, thereby leveraging both abstract rules and concrete experiences to enhance decision-making without gradient updates.

\paragraph{TRAD}

TRAD (Thought Retrieval and Aligned Decision) \cite{zhou2024trad} is a novel framework designed to enhance Large Language Model (LLM) agents in sequential decision-making tasks by addressing the limitations of traditional trajectory-level retrieval, such as irrelevant context and context window constraints. The framework comprises two core components: \textit{Thought Retrieval} and \textit{Aligned Decision}. The former employs a step-wise retrieval mechanism where the agent generates a ``thought''—an abstraction of the current state—to query a memory of expert demonstration steps, ensuring high relevance and minimizing noise. The latter augments these retrieved steps with their temporal neighbors through techniques including Temporal Expansion, Relative Order Marks, and History Alignment, thereby recovering essential contextual dynamics lost in single-step retrieval. Empirical evaluations on benchmarks such as ALFWorld and Mind2Web demonstrate that TRAD significantly outperforms state-of-the-art baselines by effectively balancing context sufficiency with noise reduction.

\subsection{Task Preprocess Details}
\label{appendix_process}

This section details the task preprocessing pipeline. As illustrated in Algorithm \ref{alg:decomposition}, raw trajectories are transformed into a hierarchical tree structure $\Psi$ through a recursive coarse-to-fine decomposition strategy. 
The process initiates with trajectory pruning, which eliminates redundant noise from the raw sequence. 
Subsequently, the sequence undergoes temporal decomposition to identify distinct sub-goals, followed by a consistency verification step where segments are validated and labeled to ensure logical coherence. 
This decomposition is applied recursively: segments containing three or more actions are subject to further partitioning, whereas shorter segments are preserved as atomic leaf nodes, thereby yielding a structured representation of long-horizon tasks. A raw trajectory is shown in Figure \ref{fig:raw task example}, and a processed example is shown in \ref{fig:processed example}.

\begin{algorithm}[htbp]
\caption{Hierarchical Task Trajectory Decomposition}
\label{alg:decomposition}
\begin{algorithmic}[1]
\Require Full task trajectory $\mathcal{T}_{\text{raw}}$
\Ensure Hierarchical Task Tree $\Psi$

\State \textbf{Initialization:} 
\State $\mathcal{T}_{\text{clean}} \leftarrow \mathcal{A}_{\text{prune}}(\mathcal{T}_{\text{raw}})$ 
\State Initialize root node $n_{\text{root}}$ with $\mathcal{T}_{\text{clean}}$
\State $\Psi.\text{setRoot}(n_{\text{root}})$
\State \Call{RecursiveDecompose}{$n_{\text{root}}$}
\State \Return $\Psi$

\vspace{0.2cm}
\Function{RecursiveDecompose}{$n_{\text{parent}}$}
    \State $\tau \leftarrow n_{\text{parent}}.\text{trajectory}$
    \State $valid \leftarrow \text{False}$
    
    \While{\textbf{not} $valid$} 
        \State $\mathcal{S} \leftarrow \mathcal{A}_{\text{split}}(\tau)$
    \EndWhile
    \ForAll{subtask $s_i \in \mathcal{S}$, label $l_i \in \mathcal{L}$}
        \State Create node $n_{\text{child}}$ with content $s_i$ and label $l_i$
        \State $\Psi.\text{addChild}(n_{\text{parent}}, n_{\text{child}})$
        
        \If{$\text{Length}(s_i) \geq 3$} 
        \Else
        \EndIf
    \EndFor  
\EndFunction
\end{algorithmic}
\end{algorithm}

\begin{figure*}
\begin{nicejson}{Raw Task Example}
Raw Trajectory 
STEP 1: go to cabinet 2 > You arrive at cabinet 2. The cabinet 2 is closed.
STEP 2: open cabinet 2 > You open the cabinet 2. The cabinet 2 is open. In it, you see a bowl 1.
STEP 3: take bowl 1 from cabinet 2 > You pick up the bowl 1 from the cabinet 2.
STEP 4: go to diningtable 1 > You arrive at diningtable 1. On the diningtable 1, you see a pen 1, a potato 1, and a tomato 1.
STEP 5: move bowl 1 to diningtable 1 > You move the bowl 1 to the diningtable 1.
STEP 6: take bowl 1 from diningtable 1 > You pick up the bowl 1 from the diningtable 1.
STEP 7: go to fridge 1 > You arrive at fridge 1. The fridge 1 is closed.
STEP 8: open fridge 1 > You open the fridge 1. The fridge 1 is open. In it, you see nothing.
STEP 9: cool bowl 1 with fridge 1 > You cool the bowl 1 using the fridge 1.
STEP 10: go to diningtable 1 > You arrive at diningtable 1. On the diningtable 1, you see a pen 1, a potato 1, and a tomato 1.
STEP 11: move bowl 1 to diningtable 1 > You move the bowl 1 to the diningtable 1.
\end{nicejson}
\caption{A task trajectory example in ALFWorld.}
\label{fig:raw task example}
\end{figure*}

\begin{figure*}
\begin{nicejson}{Example After Task Process }
|-- name: Root Task
|-- goal: put a cool bowl in diningtable. 
|-- steps_count: 11
|-- trajectory:
|-- subtasks:
|   |-- Subtask:
|   |   |-- name: Take bowl from cabinet 
|   |   |-- goal: put a cool bowl in diningtable. 
|   |   |-- steps_count: 3
|   |-- Subtask:
|   |   |-- name: Place bowl on dining table 
|   |   |-- goal: put a cool bowl in diningtable. 
|   |   |-- steps_count: 2
|   |-- Subtask:
|   |   |-- name: Cool bowl in fridge 
|   |   |-- goal: put a cool bowl in diningtable. 
|   |   |-- steps_count: 4
|   |   |   |-- Subtask:
|   |   |   |   |-- name: Take bowl from diningtable 
|   |   |   |   |-- goal: put a cool bowl in diningtable.
|   |   |   |   |-- steps_count: 1
|   |   |   |-- Subtask:
|   |   |       |-- name: Cool bowl using fridge 
|   |   |       |-- goal: put a cool bowl in diningtable.
|   |   |       |-- steps_count: 3
|   |-- Subtask:
|       |-- name: Place cooled bowl on dining table 
|       |-- goal: put a cool bowl in diningtable.
|       |-- steps_count: 2
\end{nicejson}
\caption{A processed example in ALFWorld.}
\label{fig:processed example}
\end{figure*}

\subsection{Examples of Structured Knowledge}
\label{appendix_structure_knowledge}

In this section, we provide some structured knowledge examples derived from MCMA (tree structure, chain structure, key-value structure, natural language structure, and nest structure. As shown in Figure \ref{fig:tree}, \ref{fig:chain}, \ref{fig:kv}, \ref{fig:nlp}, \ref{fig:nest}). Figure \ref{fig:generate_know_prompt} shows the prompt used for guiding memory copilot to generate structural knowledge, which outlines the role and instructions, the required JSON structures and the few-shot example provided to the model.

\subsection{Task Prompt}

Figure \ref{fig:task_prompt} shows our task prompt for ALFWorld and ScienceWorld.

\begin{figure*}
\begin{nicejson}{Task Prompt}
### HISTORY OF ACTIONS ###
Here is the history of your actions and observations so far:
{history_str}

### GOAL ###
Your ultimate goal is:
{game_task}

### CURRENT STATE ###
{current_state}

### AVAILABLE ACTIONS ###
You MUST choose one of the following valid actions. Do not make up your own actions.
Admissible actions: {admissible_commands}

Based on all the information, first think step-by-step about what to do next. Then, output your final chosen action in the format 'Action: <action>'.
\end{nicejson}
\caption{Task Prompt for our tasks.}
\label{fig:task_prompt}
\end{figure*}

\subsection{Reuse Memory Copilot on Same Series LLMs}

Since the data acquisition for each memory copilot is driven by the specific performance of the task model, the copilot is inherently aligned with the model's capabilities, thereby providing critical assistance in challenging scenarios. However, training a dedicated copilot entails significant computational overhead. To address this, we investigate the transferability of the memory copilot across different models within the same lineage. As evidenced in Tables \ref{tab:reuse4b} and \ref{tab:reuse72b}, the memory copilot optimized for Qwen3-32B demonstrates robust performance when deployed on other LLMs in the same series.

\begin{table}[htbp]
    \centering
    \begin{tabular}{ccll}
        \toprule
        \multirow{2}{*}{\textbf{Type}} & \multirow{2}{*}{\textbf{Config}} & \multicolumn{2}{c}{\textbf{Performance}} \\
        \cmidrule(lr){3-4}
        & & \textbf{Step} $\downarrow$ & \textbf{Acc} $\uparrow$ \\
        \midrule
        \textbf{Base} & \textbf{Default} & 38.67 & 0.40 \\
        \midrule
        \multirow{2}{*}{\textbf{Sum}} & \textbf{Default} & 31.39 \downgreen{7.28} & 0.51 \upgreen{11\%} \\
        & \textbf{Trained} & 31.31 \downgreen{7.36} & 0.55 \upgreen{15\%} \\
        \midrule
        \multirow{2}{*}{\textbf{Reflection}} & \textbf{Default} & 35.18 \downgreen{3.49} & 0.44 \upgreen{4\%} \\
        & \textbf{Trained} & 32.99 \downgreen{5.68} & 0.51 \upgreen{11\%} \\
        \midrule
        \multirow{2}{*}{\textbf{MCMA}} & \textbf{Default} & 30.08 \downgreen{8.59} & 0.54 \upgreen{14\%} \\
        & \textbf{Trained} & \textbf{28.37} \downgreen{10.30} & \textbf{0.59} \upgreen{19\%} \\
        \bottomrule
    \end{tabular}
    \caption{Reuse memory copilot trained for Qwen3-32B on Qwen3-4B.}
    \label{tab:reuse4b}
\end{table}

\begin{table}[htbp]
    \centering
    \begin{tabular}{ccll}
        \toprule
        \multirow{2}{*}{\textbf{Type}} & \multirow{2}{*}{\textbf{Config}} & \multicolumn{2}{c}{\textbf{Performance}} \\
        \cmidrule(lr){3-4}
        & & \textbf{Step} $\downarrow$ & \textbf{Acc} $\uparrow$ \\
        \midrule
        \textbf{Base} & \textbf{Default} & 27.93 & 0.62 \\
        \midrule
        \multirow{2}{*}{\textbf{Sum}} & \textbf{Default} & 24.68 \downgreen{3.25} & 0.66 \upgreen{4\%} \\
        & \textbf{Trained} & \textbf{22.51} \downgreen{5.42} & 0.69 \upgreen{7\%} \\
        \midrule
        \multirow{2}{*}{\textbf{Reflection}} & \textbf{Default} & 27.35 \downgreen{0.58} & 0.65 \upgreen{3\%} \\
        & \textbf{Trained} & 25.16 \downgreen{2.77} & \textbf{0.71} \upgreen{9\%} \\
        \midrule
        \multirow{2}{*}{\textbf{MCMA}} & \textbf{Default} & 24.25 \downgreen{3.68} & 0.66 \upgreen{4\%} \\
        & \textbf{Trained} & 23.70 \downgreen{4.23} & 0.69 \upgreen{7\%} \\
        \bottomrule
    \end{tabular}
    \caption{Reuse memory copilot trained for Qwen3-32B on Qwen2.5-72B.}
    \label{tab:reuse72b}
\end{table}

\subsection{Hierarchical Abstraction Processing}

This section details our approach to handling hierarchical knowledge. The process begins with \textbf{semantic embedding generation}, where we encode each knowledge entry by concatenating its name and textual description into a unified semantic vector. To model the relationships between these entries, we proceed with \textbf{sparse semantic graph construction}. Specifically, we compute the cosine similarity between knowledge vectors and construct a $k$-Nearest Neighbor ($k$-NN) graph, retaining only the top-$k$ ($k=10$) strongest connections to form a sparse adjacency matrix.

Building upon this topology, we employ \textbf{hierarchical clustering} to identify latent semantic communities within the graph. The final consolidation is achieved through a two-phase strategy: first, \textbf{intra-cluster fusion} merges redundant or highly similar knowledge points within each identified cluster; subsequently, \textbf{inter-cluster abstraction} re-aggregates these fused clusters to derive higher-level abstract concepts, thereby forming a structured and concise knowledge hierarchy.

As shown in Figure \ref{fig:abs_level1} and Figure \ref{fig:abs_level2}, we provide the examples of different level of knowledge provided by MCMA. Lower-level knowledge focuses on complete task details, while higher-level knowledge focuses on capturing common problems at higher levels.

\begin{figure*}
\begin{nicejson}{Generate Knowledge Prompt}
# ROLE AND GOAL
You are an advanced AI assistant specializing in induction and reasoning from multiple task execution trajectories. Your core objective is to identify and extract common patterns, strategies, or knowledge embedded within these trajectories and to summarize this shared knowledge in a flexible, structured format. You are required to utilize various structures, including trees, chains, graphs, and natural language to represent the inherent logic of the knowledge in the most effective way.

# INSTRUCTIONS
1. Analyze Inputs Comprehensively: Scrutinize all provided task trajectories. Each trajectory contains an overall goal and a detailed sequence of action steps.
2. Identify Common Patterns: Look for commonalities across different trajectories. These commonalities might manifest as:
    (1) Task Decomposition Structure: Is a high-level task consistently broken down into similar sub-tasks? (e.g., "place two items" is decomposed into two "place single item" sub-tasks).
    (2) Action Sequence: Does completing a sub-task follow a recurring sequence of actions? (e.g., `navigate to item -> pick up item -> navigate to destination -> place item`).
    (3) Object & Location Relationships: Is there general knowledge about where certain types of objects are typically found or should be placed?
3.  Select the Optimal Structure: For each piece of summarized common knowledge, choose the most appropriate structure for its representation:
    (1) Tree: Use to represent hierarchical relationships, such as task decomposition, where a main goal branches into sub-tasks or steps.
    (2) Chain: Use to represent strictly ordered sequences of actions or processes.
    (3) Key-Value: Use to represent properties, attributes, or mappings between entities (e.g., an object and its common locations).
    (4) Natural Language: Use to provide an intuitive, human-readable summary of complex patterns or insights.
4.  Combine and Nest Structures: You have the flexibility to combine and nest these structures. For example, a leaf node in a tree structure could itself be a chain structure detailing the execution flow for that sub-task.
5.  Format the Output: Consolidate all extracted knowledge into a single JSON object. This object must contain a list named `knowledge`, where each element represents a distinct piece of knowledge and adheres to the JSON structures defined below.

# STRUCTURE FORMATS
...

# EXAMPLE OUTPUT:
Answer: {
  "General Plan for Placing Two Items in a Receptacle": {
    "name": "General Plan for Placing Two Items in a Receptacle",
    "structured_storage": {
    "type": "tree",
    "root": {"name": "put two <item> in <receptacle>",
        "children": [{
            "name": "Process first <item>",
            "children": [{ "structured_storage": {"type": "chain","nodes": [{"step": "go to location of <item> 1"},{"step": "take <item> 1"},{"step": "go to <receptacle>"},{"step": "(optional) open <receptacle> if it is closed"},{"step": "move <item> 1 to <receptacle>"},...
]}}]}}

# TASK START
Analyze the following input data and generate the JSON output according to the formats instructed. Start with 'Answer:' before outputting the official answer.
[input_data]
\end{nicejson}
\caption{The prompt to guide memory copilot in generating structured knowledge.}
\label{fig:generate_know_prompt}
\end{figure*}

\begin{figure*}
\begin{nicejson}{Tree Example}
{
  "name": "Inspect an Object Using a Light Source",
  "description": "A process for examining a specific <object> using a <light source>. It begins by navigating to the location of the <light source> and turning it on, then proceeds to locate and retrieve the <object> from its <location>, and finally returns to the <light source> location to inspect the <object> under the illumination.",
  "knowledge": {
    "name": "General Procedure for Observing an Object Under a Light Source",
    "source": ["look at cellphone under the desklamp"],
    "structured_storage": {
      "type": "tree",
      "root": {
        "name": "look at <object> under <light source>",
        "children": [
          { "name": "Navigate to <light source> location",
            "children": [{ "structured_storage": { "type": "chain", "nodes": [
                    { "step": "go to location of <light source>" },
                    { "step": "use <light source> to turn it on" }]}}]},
          { "name": "Locate and Retrieve <object>",
            "children": [{ "structured_storage": { "type": "chain", "nodes": [
                    { "step": "go to location of <object>" },
                    { "step": "take <object> from <location>" }]}}]},
          { "name": "Return to <light source> location",
            "children": [{ "structured_storage": { "type": "chain", "nodes": [
                    { "step": "go back to location of <light source>" }]}}]}
]}}}}
\end{nicejson}
\caption{Tree structure knowledge examples.}
\label{fig:tree}
\end{figure*}

\begin{figure*}
\begin{nicejson}{Chain Example}
{
  "name": "Retrieve an Object with Lighting Assistance",
  "description": "A process for retrieving an object using a light source for assistance. It begins by navigating to the light source and turning it on, then moving to the object's location to retrieve it, and finally returning to the light source.",
  "knowledge": {
    "name": "Core Action Sequence for Observing an Object Under a Light Source",
    "source": ["look at cellphone under the desklamp"],
    "structured_storage": {
      "type": "chain",
      "nodes": [
        { "step": "Navigate to the location of the light source." },
        { "step": "Turn on the light source." },
        { "step": "Navigate to the location of the object." },
        { "step": "Retrieve the object from its location." },
        { "step": "Return to the location of the light source." }
      ]
    }
  }
}
\end{nicejson}
\caption{Chain structure knowledge examples.}
\label{fig:chain}
\end{figure*}

\begin{figure*}[h]
\begin{nicejson}{Key-Value Example}
{ 
  "name": "Organize Items in Designated Locations",
  "description": "A simple process for placing specific items in their designated receptacles. The task involves identifying an <item> and its corresponding <receptacle> from a predefined mapping, and then moving the <item> to the specified <receptacle> for organization.",
  "knowledge": {
    "name": "Common Object Locations",
    "source": ["look at cellphone under the desklamp"],
    "structured_storage": {
      "type": "key_value",
      "data": { "cellphone": ["armchair"], "desklamp": ["desk"] 
      }
    }
  }
}
\end{nicejson}
\caption{Key-Value structure knowledge examples.}
\label{fig:kv}
\end{figure*}

\begin{figure*}
\begin{nicejson}{Natural Language Example}
{ 
  "name": "Observe an Object Under a Light Source",
  "description": "A structured process where the agent navigates to a light source, activates it, locates and retrieves a target object, and then returns to the light source to observe the object under its illumination.",
  "knowledge": {
    "name": "Iterative Task Strategy for Observing an Object Under a Light Source",
    "source": ["look at cellphone under the desklamp"],
    "structured_storage": {
      "type": "natural_language",
      "text": "The agent follows a structured, iterative approach to observe an object under a light source. It first navigates to the light source, turns it on, then locates and retrieves the object, and finally returns to the light source to observe it under the light."
    }
  }
}
\end{nicejson}
\caption{Natural Language structure knowledge examples.}
\label{fig:nlp}
\end{figure*}

\vspace{-20pt}
\begin{figure*}
\begin{nicejson}{Nested Example (Tree + Chain)}
{"name": "Inspect an Object Using a Light Source",
  "description": "A process for examining a specific <object> using a <light source>.",
  "knowledge": {
    "name": "General Procedure for Observing an Object Under a Light Source",
    "source": ["look at cellphone under the desklamp"],
    "structured_storage": {
      "type": "tree",
      "root": {
        "name": "look at <object> under <light source>",
        "children": [
          { "name": "Navigate to <light source> location",
            "children": [{ "structured_storage": { "type": "chain", "nodes": [
                    { "step": "go to location of <light source>" },
                    { "step": "use <light source> to turn it on" }]}}]},
          { "name": "Locate and Retrieve <object>",
            "children": [{ "structured_storage": { "type": "chain", "nodes": [
                    { "step": "go to location of <object>" },
                    { "step": "take <object> from <location>" }]}}]},
          { "name": "Return to <light source> location",
            "children": [{ "structured_storage": { "type": "chain", "nodes": [
                    { "step": "go back to location of <light source>" }]}}]}]}
    }
  }
}
\end{nicejson}
\caption{Nested structure knowledge examples.}
\label{fig:nest}
\end{figure*}

\begin{figure*}[h]
\begin{nicejson}{Knowledge of Abstract Level 1 Example}
{
  "name": "Cool and Replenish Mug in Coffee Machine",
  "description": "An iterative process for preparing a cooled mug for use in a coffee machine. The task begins by retrieving the mug from a <container> (e.g., a cabinet), moving it to the coffee machine, and then returning it to cool in a <cooling_receptacle> (e.g., a fridge). Once cooled, the mug is placed back into the coffee machine for use.",
  "knowledge": {
    "name": "Iterative Task Strategy for Mug Placement",
    "source": ["put a cool mug in coffeemachine"],
    "structured_storage": {
      "type": "natural_language",
      "text": "The agent follows an iterative process for placing a cool mug in the coffee machine. It first retrieves the mug from the cabinet, moves it to the coffee machine, takes it back, cools it in the fridge, and then places the cooled mug back into the coffee machine."
    }
  }
}
\end{nicejson}
\caption{An Example Knowledge of Abstract Level 1.}
\label{fig:abs_level1}
\end{figure*}

\begin{figure*}
\begin{nicejson}{Knowledge of Abstract Level 2 Example}
{
  "name": "General Plan for Cooling and Storing an Item",
  "description": "A generic, multi-step process for cooling and storing an object. The task begins with retrieving a specific <item> from its <container>, proceeds to cool the <item> using a <cooling_source>, and concludes by placing the cooled <item> into a designated final <receptacle> for storage.",
  "structured_storage": {
    "type": "tree",
    "root": {
      "name": "cool <item> and put it in <receptacle>",
      "children": [
        { "name": "Retrieve <item>",
          "children": [{ "structured_storage": { "type": "chain", "nodes": [{ "step": "go to location of <item>" }, { "step": "open <container> if it is closed" },{ "step": "take <item> from <container>" } ]}}]},
        { "name": "Cool <item>",
          "children": [{ "structured_storage": { "type": "chain", "nodes": [{ "step": "go to <cooling_source>" }, { "step": "open <cooling_source> if it is closed" }, { "step": "cool <item> with <cooling_source>" } ]}}]},
        { "name": "Place <item> in <receptacle>",
          "children": [{ "structured_storage": { "type": "chain", "nodes": [{ "step": "go to <receptacle>" }, { "step": "move <item> to <receptacle>" } ]}}]}
      ]
    }
  }
}
\end{nicejson}
\caption{An Example Knowledge of Abstract Level 2.}
\label{fig:abs_level2}
\end{figure*}

\end{document}